%% file: acl_latex.tex
\pdfoutput=1
\documentclass[11pt]{article}

\usepackage{acl}
\usepackage{times}
\usepackage{latexsym}
\usepackage[T1]{fontenc}
\usepackage[utf8]{inputenc}
\usepackage{microtype}
\usepackage{inconsolata}
\usepackage{graphicx}
\usepackage{booktabs}
\usepackage{tabularx}

\usepackage{algorithm}
\usepackage{algorithmic}
\usepackage{amsmath}
\floatname{algorithm}{Technique}
\usepackage{enumitem}
\usepackage{titlesec}

\usepackage{listings}
\usepackage{xcolor}
\lstdefinelanguage{json}{
    basicstyle=\ttfamily\small,
    numbers=left,
    numberstyle=\tiny\color{gray},
    stepnumber=1,
    numbersep=5pt,
    showstringspaces=false,
    breaklines=true,
    frame=lines,
    backgroundcolor=\color{lightgray!20},
    literate=
     *{0}{{{\color{blue}0}}}{1}
      {1}{{{\color{blue}1}}}{1}
      {2}{{{\color{blue}2}}}{1}
      {3}{{{\color{blue}3}}}{1}
      {4}{{{\color{blue}4}}}{1}
      {5}{{{\color{blue}5}}}{1}
      {6}{{{\color{blue}6}}}{1}
      {7}{{{\color{blue}7}}}{1}
      {8}{{{\color{blue}8}}}{1}
      {9}{{{\color{blue}9}}}{1}
}

\title{On the Temporal Question-Answering Capabilities of Large Language Models Over Anonymized Data}

\author{Alfredo Garrachón Ruiz \and Tomás de la Rosa \and Daniel Borrajo \\
        AI Research, JP Morgan Chase}

\begin{document}
\maketitle
\begin{abstract}
The applicability of Large Language Models (LLMs) in temporal reasoning tasks over data that is not present during training is still a field that remains to be explored. In this paper we work on this topic, focusing on structured and semi-structured anonymized data. We not only develop a direct LLM pipeline, but also compare various methodologies and conduct an in-depth analysis. We identified and examined seventeen common temporal reasoning tasks in natural language, focusing on their algorithmic components. To assess LLM performance, we created the \textit{Reasoning and Answering Temporal Ability} dataset (RATA), featuring semi-structured anonymized data to ensure reliance on reasoning rather than on prior knowledge. We compared several methodologies, involving SoTA techniques such as Tree-of-Thought, self-reflexion and code execution, tuned specifically for this scenario. Our results suggest that achieving scalable and reliable solutions requires more than just standalone LLMs, highlighting the need for integrated approaches.


\end{abstract}

\input{Sections/1.introduction}

\input{Sections/2.related_work}

\input{Sections/3.background}

\input{Sections/4.temporal_tasks}

\input{Sections/5.reasoning_techniques}

\input{Sections/6.datasets}

\input{Sections/7.evaluation}

\input{Sections/8.aplicability}

\input{Sections/9.temporal_condifence}

\input{Sections/10.discussion}

\input{Sections/11.conclusion}

\input{Sections/13.limitations}

\input{Sections/12.disclaimer}

\bibliography{references}
\appendix
\clearpage
\section{Supplementary Material}

\subsection{Rest of Techniques} \label{rest_techniques}

\subsubsection{Chain of Thought (CoT)} 

Following the philosophy of divide and conquer~\cite{wei2022chain}, we propose using this technique to solve TQA tasks with a 3-prompt sequence.

\begin{algorithm}
\caption{Chain of Thought (CoT)}
\begin{algorithmic}[1]
\STATE \textbf{Prompt 1:} Analyze the TQA task and the given data (in semi-structured natural language), think about the best way to solve it.
\STATE \textbf{Prompt 2:} Represent the data in the format you deem most appropriate.
\STATE \textbf{Prompt 3:} Solve the TQA task.
\end{algorithmic}
\end{algorithm}

\subsubsection{Tree of Thought (ToT)}
The tree of thought (ToT) aims to not limit the reasoning process to a predefined path, but to let the LLM reason over different self-defined paths and find the most suitable one~\cite{yao2024tree}. In this implementation, we let the LLM create different reasoning paths up to level 2 of depth, with the LLM itself acting as a judge and a beam width of 2.
\begin{algorithm}
\caption{Tree of Thought (ToT)}
\begin{algorithmic}[1]
\STATE \textbf{Initial prompt:} Introduction about the ToT, the semi-structured data in natural langugage and the lask to be carried out.
\FOR{each depth level}
    \FOR{each node in the frontier}
        \STATE \textbf{Expand:} Generate 3 reasoning steps.
        \STATE \textbf{Evaluate:} Assign scores to each state with backtracking reasoning path.
    \ENDFOR
    \STATE \textbf{Select:} Retain top nodes at frontier based on beam width.
\ENDFOR
\STATE \textbf{Final prompt: } Solve the TQA task based on best reasoning path found.
\end{algorithmic}
\end{algorithm}

\begin{table}[h!]
\centering
\small
\begin{tabular}{lrrr}
\toprule
  Method &  Accuracy &  Average Time &  Time (Std Dev) \\
\midrule
  Direct &      64\% &          3.18 &            2.13 \\
     CoT &      66\% &         37.57 &           55.52 \\
     ToT &      68\% &         56.40 &          215.43 \\
    CoTR &      76\% &         35.07 &          159.29 \\
  CoTE-S &      72\% &        111.76 &          185.36 \\
    CoTE &      84\% &         13.09 &            9.09 \\
CoTAPI-S &      86\% &        106.33 &          248.61 \\
  CoTAPI &      \textbf{93\%} &          8.59 &            4.70 \\
\bottomrule
\end{tabular}
\caption{Accuracy and Time values in the Test-of-Time benchmark}
\label{tab:accuracy_tot}
\end{table}

\input{Sections/14.prompts}

\input{Sections/15.functions}

\end{document}

%% file: Sections/1.introduction.tex
\section{Introduction}

Large Language Models (LLMs) have demonstrated remarkable performance across a wide range of natural language processing (NLP) tasks, making them a promising tool for enabling AI-business applications ~\cite{brown2020fewshot_llm,chung2022scalinginstructionfinetunedlanguagemodels,dubey2024llama3herdmodels}. Recent studies have focused on uncovering and enhancing the reasoning capabilities of LLMs, exploring new methodologies to make them more adept at complex cognitive tasks~\cite{huang2023reasoninglargelanguagemodels,wei2022chain}. Among these, temporal reasoning has gained interest~\cite{wang2023tram,fatemi2024test,xiong2024learn_tr}.
Within the realm of temporal reasoning, we focus on Temporal Question Answering (TQA), a task that necessitates the accurate interpretation of events over time~\cite{su2024_TKGQA_survey}. Most studies focus on general knowledge or public data that most-likely was present during the training phase of the LLMs~\cite{wang2023tram,TempQuestions}, as well as semantic rich pieces of data~\cite{chu2023timebench} and scenarios where the answer is among a known set~\cite{su2024timobettertemporalreasoning,wang2023tram}. In this study we are going to focus in a less-explored scenario of structured and semi-structured anonymized data that the LLM has never seen previously, trying to effectively measure its capabilities of reasoning without using previous knowledge.

TQA queries often involve temporal relations (e.g., before, during, between) and time-related expressions that can be either explicit (e.g., specific years like 2024, dates, or time periods as "from 1994 to 2000") or implicit (e.g., events like "the end of World War II"). This kind of task is essential for applications in environments where the timing of events is crucial, such as financial markets or sports analytics.



In this work, we argue that to achieve scalable and reliable reasoning for TQA over structured or semi-structured anonymized data, LLMs should be leveraged not as standalone problem solvers, but as components in a broader system. Specifically, LLMs can be used effectively with in-context learning to identify the objective of the task, extract the relevant elements and select the proper solvers when the type of task is recognizable beforehand. For cases where the task type is unknown, we can keep the LLMs flexibility by letting them generate the executable logic required to solve it. The underlying idea is that the actual algorithmic operations and execution should be delegated to external tools or functions that can perform these tasks more efficiently.


The main contributions of this work are the following. First, we provide a detailed analysis of the typology of questions within TQA, which allows us to identify the low-level algorithms required for effective resolution. This analysis has led to: the discovery of 17 types of questions among the most popular ones; and the development of an API composed of external functions associated to each of the identified questions. Second, we created a new TQA dataset which comprises the 17 types of tasks previously identified with semi-structured anonymized data so the LLM cannot rely on prior training knowledge, but rather on its reasoning abilities, making this a very robust dataset for evaluating LLMs' temporal reasoning capabilities. Third, we conduct an experimental evaluation comparing various methodologies involving LLMs tailored to the TQA task over structured and semi-structured anonymized data, including Chain of Thought (CoT)~\cite{wei2022chain}, Tree of Thought (ToT)~\cite{yao2024tree}, Chain of Thought with Reflexion (where the LLM can evaluate and retry the task in a more guided approach \cite{shinn2023reflexion}), CoT with External Execution (where the LLM generates algorithm-related code), and CoT with API Access (where the LLM can call upon the previously developed functions). Finally, we offer a deep analysis and discussion of the results, with the extra added value of checking the applicability of the techniques in different scenarios, as well as the the confidence of the LLM when handling a new task with the goal of its embodiment in an TQA agent.

%% file: Sections/2.related_work.tex
\section{Related Work}
In this section we summarize the advances of temporal reasoning capabilities of language models, and how they relate to previous approaches of temporal reasoning within AI.

\subsection{Temporal Reasoning} The main approach to answer time-related questions is to reason over a  temporal knowledge graph (TKG), which is a knowledge graph annotated with time intervals and/or temporal constraints~\cite{su2024_TKGQA_survey,cai2024_TKG_rep_survey}. Various TKG embedding are used to match queries and relevant facts~\cite{mavromatis2022tempoqr, liu2023timeawaremultiwayadaptivefusion}, so the resulting sub-graph can be analyzed considering the time constraints in order to rank candidate answers~\cite{chen2022temporal}. However, these methods require specialized modules to train and evaluate time-aware embeddings.

\subsection{Temporal Reasoning with LLM} 
Several studies evaluate the LLMs ability to answer TQA directly, but the intention of these evaluations is to release benchmarks~\cite{wang2023tram,fatemi2024test}, or to set a baseline against which other techniques try to improve direct answering skills~\cite{tan2023towards}.
To enhance temporal reasoning capabilities of LLMs various \textit{in-context learning}~\cite{brown2020fewshot_llm} and Chain-of-Thought~\cite{wei2022chain} approaches have been proposed. ARI~\cite{chen-etal-2024-temporal} divides the task in knowledge-base and knowledge-agnostic (where the temporal reasoning itself happens guided by the LLM) and TG-LLM~\cite{xiong2024learn_tr} split the reasoning in (1) translating the task in natural language into a TKG, and (2) reasoning over the TKG with specialized CoT per question type.
These techniques are attractive due to the flexibility with respect to the input. Nevertheless, they can not provide guarantees regarding the temporal related algorithms as the reasoning steps still operate at the natural language level.

\subsection{LLM-generated Programs} An alternative to directly reasoning at the natural language level is to ask LLMs to come up with the problem-solving procedure in an intermediate logic/algorithmic form~\cite{nie2024codeQA, chen2024selfimprovementprogrammingtemporalknowledge}, or directly in a program written in a popular programming language~\cite{wang2024executable}. Then, the execution of this program provides an answer that can be easily incorporated into an NL-like response. These methods are targeting general question-answering, so the potential for TQA is still an open research line. 

\subsection{Reasoning with Solvers/External APIs} To provide precise and reliable problem solving skills, LLMs have been complemented with access to external solvers such as automated planners~\cite{liu2023llm+plan}, constraint satisfaction and boolean satisfiability problem (SAT) solvers~\cite{pan2023logic}. Following this spirit, the idea of our work is to provide access to an API with functions that will solve TQA tasks.


%% file: Sections/3.background.tex
\section{Background}
A temporal knowledge graph (TKG) is a directed multi-relational graph $\mathcal{G}= (E,R, K)$, where $E$ is a set of entities (nodes), $R$ is the set of relation names, and $K$ is the set of edges, representing temporal annotated facts. Each fact $k \in K$ is represented as $(s,r,o,t_0, t_1)$, where $s, o \in E$ are the fact subject and object, $r$ is the relation, and $[t_0, t_1]$ is the interval when the relation holds. For instance, the fact \textit{("Barack Obama","president", "USA", 2013, 2017)} represents Obama's second term as USA president. Time intervals can be expressed with timestamps of any level of granularity. Without loss of generality we will express the examples in years. 

In our TQA context, a question in natural language is parsed to produce a logic query $q$ (or a set of queries) to $\mathcal{G}$ to retrieve a set $F$ of relevant facts from which an answer can be generated. $F$ may be represented as a sub-graph of $G$ on convenience \cite{chen2022temporal}. Part of this semantic parsing implies the identification of the elements to consider. Thus, $q(s,?,?)$ cares about the subject, $q(?,r,?)$ about the relation, $q(s,r,?)$ about the relations of the subject, and so on. Following this logic "\textit{q(?, president, USA)} will retrieve the facts stating the terms of all USA presidents. 

%% file: Sections/4.temporal_tasks.tex
\section{Temporal Tasks} \label{temporal_tasks}

In TQA, retrieving the facts related to the entities mentioned in the question is not enough. In order to perform the task, in most scenarios it is required to carry on some operations based on the associated temporal data. Regardless of the classification of tasks in TQA, each task can be identified by the sequence of these low-level algorithms required to solve the task.
 

These low-level algorithms are shared among many temporal reasoning tasks, allowing for the categorization of different tasks based on the combinations of these. Apart from the various combinations of logic queries to $\mathcal{G}$, we have identified a set of common relevant key primitive algorithms that underpin most TQA tasks. These are:

\begin{itemize}[itemsep=2pt, topsep=3pt]
    \item \textbf{Sort:} Arrange different facts in chronological order. For a set $F'$ of (relevant) facts, sort them by $t_0$ (also possible by $t_1$). For example, \textit{"Which was the first team in which Henry played football?"} requires to: (1) create the query \textit{q(Henry, player of, ?)}; (2) sort all retrieved facts by $t_0$; and (3) return the object from the first fact of the sorted list.
    \item \textbf{Count:} Determine the number of relationships or entities. For a set $F'$ of facts, count the number of edges that meet a specific condition. For example, \textit{"How many times has Nadal won Roland Garros in the first 15 years of this century?"} requires to: (1) create the query \textit{q(Nadal, winner of, Roland Garros)}; (2) filter the results to facts ($t_0$ and $t_1$) between 2000 and 2015; and (3) count the filtered facts.
    \item \textbf{Filter Time:} Select facts based on specific time points. For a set $F'$ of facts, retrieve only those facts that meet the specified date criteria. For example, \textit{"Who was the champion of the Champions League in 2014?"} requires to: (1) create the query \textit{q(?, champion of, Champions League)}; and (2) filter the results to the fact where $t_0 <= 2014 <= t_1$.
    \item \textbf{Interval Duration:} Filter facts based on the duration of time intervals. For a set $F'$ of facts, perform operations based on the time interval $[t_0, t_1]$ of each one. For example, \textit{"During Cold War, who were the presidents of the US that lasted longer than 4 years?"} requires to: (1) create the query \textit{q(?, president of, USA)}; (2) filter the results to include only facts where the interval $[t_0, t_1]$ overlaps with 1947–1991; and (3) filter the facts where $(t_1 - t_0) >= 4$.
\end{itemize}

In order to provide an analysis of the use of these primitive functions for different TQA tasks, we have both reviewed the literature, and reasoned through the various temporal reasoning tasks that frequently occur in natural human language. Each of these identified tasks has been decomposed into the different previously defined primitive algorithms required. Details are depicted in Table~\ref{tab:question_types}.

\begin{table*}[h!]
\centering
\scriptsize
\begin{tabular}{| m{3.5cm} | m{0.4cm} | m{0.6cm} | m{0.5cm} | m{0.9cm} | m{5.5cm} | m{1.3cm} |}
\hline
\textbf{Task} & \textbf{Sort} & \textbf{Filter \newline Time} & \textbf{Count} & \textbf{Interval \newline Duration} & \textbf{Example} & \textbf{Dataset} \\
\hline
\hline 
Sort entities & \textbf{X} & & & & List, in a sorted way, all the presidents of the US & ToT, TQ\\
\hline
Before after & \textbf{X} & & & & Where Thomas lived before living in Stockholm & ToT, CQ, TQ\\
\hline
Entity at specific time & & \textbf{X} & & & Who was the coach of New York Yankees in 2020 & ToT, CQ, TQ\\
\hline
Event at what time & & & & & When Argentina ceased to be a Spanish colony & ToT, CQ, TQ\\
\hline
First last & \textbf{X} & & & & Who was the first CEO of OpenAI & ToT, CQ, TQ\\
\hline
Event at the time of another event & & \textbf{X} & & & When Messi started playing football, for which team was Ronaldo playing for & ToT, CQ, TQ\\
\hline
Count events in time period & & & \textbf{X} & & How many countries joined the OTAN from 1980 to 2010 & ToT\\
\hline
Duration of event & \textbf{X} & & & \textbf{X} & How long was Zidane coach of Madrid the 2nd time & ToT\\
\hline
Longest or shortest event duration & \textbf{X} & & & \textbf{X} & Which video game has the time record of top 1 in sales & \\
\hline
List entities at event time & \textbf{X} & \textbf{X} & & & List the unicorn companies in Portugal while Antonio Costa was prime minister & \\
\hline
In between entities & \textbf{X} & \textbf{X} & & & Since Apple released the iPhone 7 till the release of the iPhone 14, list the rest of products released & \\
\hline
Relation total duration & & & & \textbf{X} & How long was Phil Jackson a coach & \\
\hline
Event within another Event & & \textbf{X} & & & Did the Normandy Landing occur during the WW2? & \\
\hline
Without relation & \textbf{X} & & & \textbf{X} & Was Intel without releasing a processor for more than 2 years between 2015 and 2022? & \\
\hline
Duration comparison & & & & \textbf{X} & Was Angela Merkel more time chancellor of Germany compared to Helmut Kohl? & \\
\hline
Event sequence pattern & & \textbf{X} & & & Did Google release and cease ``Stadia'' in between 2018 and 2023? & \\
\hline
Count relations with duration & & & \textbf{X} & \textbf{X} & How many interactions longer than 5 years have been between USA and Japan from 1900 and 2024? & \\
\hline
\end{tabular}
\caption{Relation of the defined primitive temporal algorithms and several reasoning tasks with their appearances in existing datasets (See Section \ref{datasets})}
\label{tab:question_types}
\end{table*}

%% file: Sections/5.reasoning_techniques.tex
\section{Reasoning Techniques} \label{reasoning_techniques}
After studying and categorizing TQA tasks, we move on to explore the applicability of LLMs in resolving them from various perspectives, performing different roles (See Figure \ref{fig:diagram1}). The more basic techniques of \textbf{Chain of Thought (CoT)} and \textbf{Tree of Thought (ToT)} are explained in the Supplementary Material (see Section~\ref{rest_techniques}). In the scenarios where the LLM can not directly access the structured data or TKG, we make use of an equivalent natural language representation of the TKG facts (semi-structured representation).

\subsection{Chain of Thought with Self-Reflexion}

\begin{algorithm}[H]
\caption{CoT with Reflexion}
\begin{algorithmic}[1]
\STATE \textbf{Prompt 1:} Analyze the TQA task and data (semi-structured), think about how to solve it.
\WHILE{not proper solution}
    \STATE \textbf{Prompt 2:} Identify and extract the most relevant data to the task.
    \STATE \textbf{Prompt 3:} Solve the TQA task.
    \STATE \textbf{Evaluate} Evaluate the answer and, if it seems not proper, reason why
\ENDWHILE
\end{algorithmic}
\end{algorithm}

In this scenario we mix the CoT with the reflexion technique with the objective to analyze the answer obtained by the LLM, and if it seems incorrect or not trustable, obtain verbal feedback that would be useful to the LLM when iterating again \cite{shinn2023reflexion}.

\subsection{Chain of Thought with External Execution}

\begin{algorithm}[H]
\caption{CoT with External Execution}
\begin{algorithmic}[1]
\STATE \textbf{Prompt 1:} Analyze the task and a subset of facts \( K' \) from the TKG to know its format.
\STATE \textbf{Prompt 2:} Generate the code to solve the task in a executable environment with access to the structured data.
\STATE \textbf{External:} Extract the code and execute it.
\end{algorithmic}
\end{algorithm}

The idea behind this technique stems from the fact that an LLM cannot actually execute the necessary algorithms. Although these models exhibit high language-oriented capabilities due to the extensive training data, their inference time per token remains constant regardless of the task's complexity, showing it will just select the next high-probability tokens learnt at training. In this context, we aim to explore whether the code generation capability of LLMs can be leveraged to achieve better results than if these TQA tasks were reasoned directly by the LLM.

\subsection{Chain of Thought with a pre-defined function set} 
In this scenario the LLM uses an external API that contains the necessary set of functions required to obtain the answer in conjunction with In-Context Learning (ICL) techniques. Each function corresponds to a task category described in Table~\ref{tab:question_types}. This approach aims to relieve the LLM of the burden of generating code by allowing it to utilize deterministic functions that have been tested and verified. Hence, the LLM focuses on analyzing the task from a semantic perspective, selecting the most appropriate function, and identifying the necessary parameters for the task resolution.

\begin{algorithm}[H]
\caption{CoT with a pre-defined function set}
\begin{algorithmic}[1]
\STATE \textbf{Prompt 1:} Analyze the task and a subset of facts \( K' \) from the TKG to know its format.
\STATE \textbf{Prompt 2:} Based on a schema of API functions with examples (ICL philosophy), identify the appropriate function to execute, along with the required parameters.
\STATE \textbf{Prompt 3:} Reason over the result to determine the answer.
\end{algorithmic}
\end{algorithm}



As an example, for the task \textit{"How long was Phil Jackson a coach"}, the LLM should identify, by the provided schema, that this type of question is "Relation Total Duration", with the associated function format of \textit{f(e, r, ?)}. This will require the LLM to identify correctly the parameters "\textit{f(Phil Jackson, coach, ?)}" to solve the task.

\subsubsection{External API}
Based on the analysis conducted in Section~\ref{temporal_tasks}, we have developed a set of functions for the resolution of each task type. Each of the developed functions is composed of three elements \textit{\{l, d, u\}}: logic, description, and use guide. The logic (\textit{l}) relates to the logic code that, given the correct parameters, solves the task and returns the result. The description (\textit{d}) is composed of the scenarios on which to use the function, the behavior of it, and what to expect as a result. This description is accompanied by a few-shot (ICL) of 1 to 2 examples to give more context to the LLM. Finally, the use guide (\textit{u}) comprises the description of the parameters required by the function, so the LLM knows what to look for. The scheme that is delivered to the LLM is composed of the description (\textit{d}) and the use guide (\textit{u}) for each of the functions of the set, while the LLM remains agnostic of the (\textit{l}) behind each function.

%% file: Sections/6.datasets.tex
\section{Datasets} \label{datasets}
We have analyzed the available relevant open source TKGQA (Temporal Knowledge Graph Question Answering) datasets such us: Test-of-time~\cite{fatemi2024test}, CronQuestions~\cite{cronquestions}, TempQuestions~\cite{TempQuestions} and more. The previously identified types of TQA tasks have been classified into these (three) datasets as can be seen in Table~\ref{tab:question_types}. For our experimentation the data should be semi-structured and anonymized or synthetic, so that the LLM does not rely on training knowledge, excluding then some datasets that use publicly available information \cite{zhang-etal-2023-crt, gupta2023temptabqatemporalquestionanswering}. This setting is adequate for evaluating scenarios where the data to be reasoned upon are private or new, such as corporate data or recent information not included in the training corpus.


Among these, we selected the Test-Of-Time (ToT)~\cite{fatemi2024test} as the most suitable one for experimentation. This comprises eight types of questions that fall within the set of 17 question types previously described (Table~\ref{tab:question_types}), and were also synthetically generated using synthetic graphs, resulting in a total of 2,800 questions.
There still remain nine types of questions that are not covered by this dataset. Therefore, we have adopted the approach of this dataset and extended it to generate new data. To this end, we have generated several synthetic graphs of different types and created questions based on these, with five to ten templates for each question type to transform them into natural language, ensuring variability. This new dataset contains 3,050 questions across nine different question types. We have combined both datasets to form the complete dataset, named \textbf{Reasoning and Answering Temporal Ability dataset - RATA}. This comprises the 17 question types most common in natural language (compared to Tot covering 8, TempQuestions covering 6 and CronQuestions covering 5) with both implicit and explicit time references for a total of 5,850 questions with their associated semi-structured anonymized data.

Additionally, we want to show the applicability of the proposed methodology to other scenarios. To this extend, the CronQuestions dataset~\cite{cronquestions} has been selected as others like the TempQuestions do not fullfil the requirements of semi-structured and anonymized data. For the evaluation we used the complex subset in its anonymized version, composed of 5,050 tasks.

We also aimed to test the LLM's ability to distinguish between TQA and KQA tasks with the idea of developing a temporal-expert LLM agent. For this, we selected the ComplexWebQuestions dataset~\cite{talmor2018webknowledgebaseansweringcomplex}. This Knowledge Base Question Answering (KBQA) dataset consists of question-answering tasks based on internet data. An excerpt of this dataset was taken and manually filtered from tasks related to temporal reasoning to create a corpus of 1,000 purely knowledge-based tasks.

%% file: Sections/7.evaluation.tex
\section{Evaluation}
The evaluation of this research comprises three distinct experiments. First, the proposed techniques will be tested on the RATA (Reasoning and Answering Temporal Ability) dataset. Secondly, the applicability of the developed techniques will be tested in a different scenario to demonstrate their universal use in related scenarios. Finally, an experiment will be conducted on the LLM's confidence in identifying temporal tasks, serving as a proof of concept for the TQA-expert agent.


In the scenarios of \textit{CoT with external execution} and \textit{CoT with a pre-defined function set}, there are two variants to effectively cover the structured and semi-structure data scenarios. In the structured data scenario the data will be already in a graph format for which the LLM actions will have access to. In the semi-structured data scenario, the TKG data (in the equivalent natural language representation) are included in the initial prompt instead of just the excerpt, same as in the other techniques. Consequently, an additional prompt is added to the chain, instructing the LLM to generate a code to transform the data into a TKG upon which the task can subsequently be solved.

\begin{figure*}[htbp]
    \centering
    \includegraphics[width=0.8\textwidth]{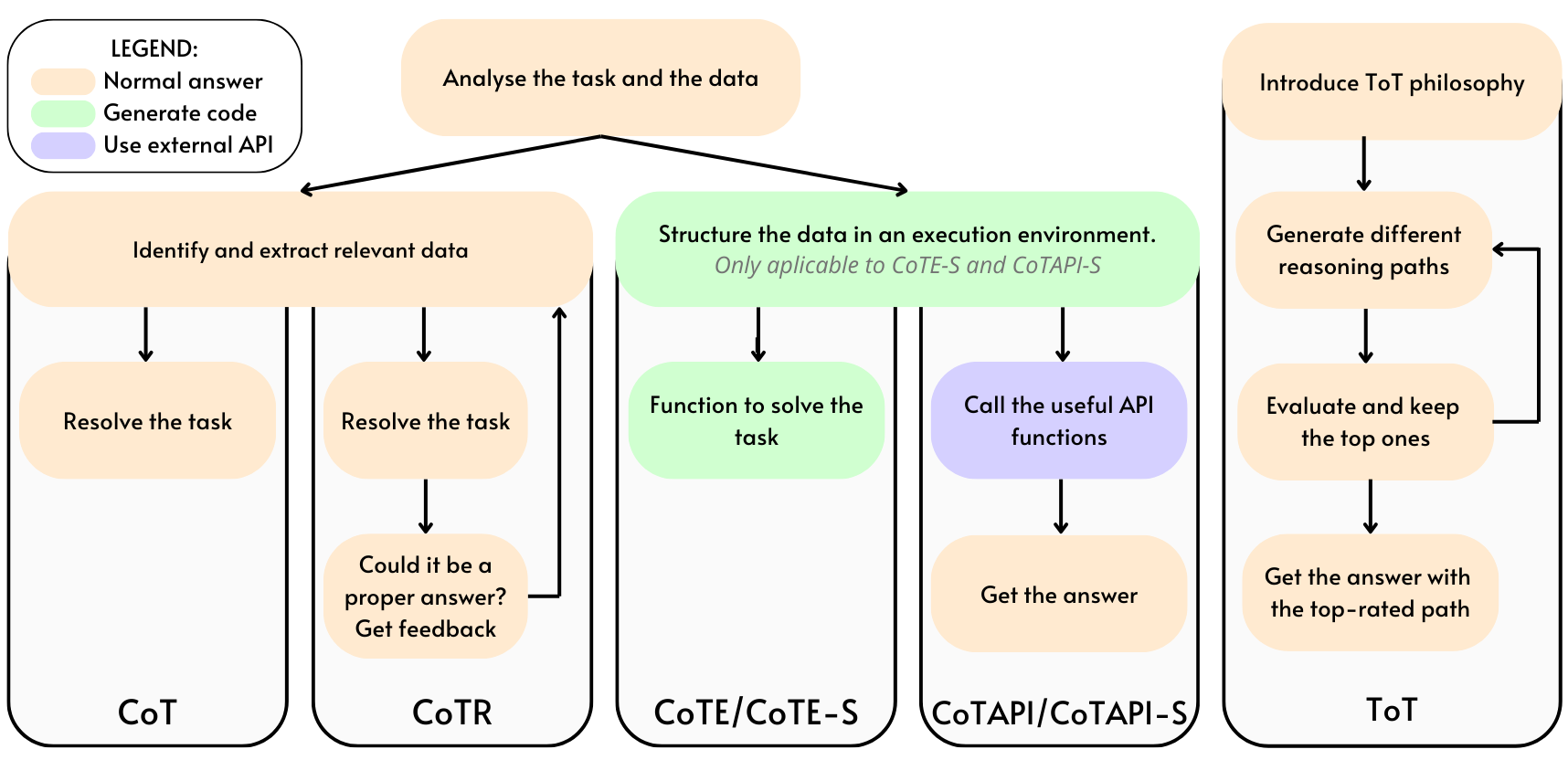}
    \caption{Diagram with the different techniques proposed.}
    \label{fig:diagram1}
\end{figure*}

In total, six different techniques will be tested and compared. Below is a summary of the nomenclature of each of these techniques, and in Figure~\ref{fig:diagram1} a diagram about them.

\begin{itemize}[noitemsep, topsep=0pt]
    \item \textbf{Direct: } Single prompt requesting the answer
    \item \textbf{CoT: } Chain of Thought
    \item \textbf{ToT: } Tree of Thought
    \item \textbf{CoTR: } Chain of Thought with reflexion
    \item \textbf{CoTE: } CoT with external execution
    \item \textbf{CoTE-S: } CoT with external execution and structuring of the data
    \item \textbf{CoTAPI: } CoT with a pre-defined function set
    \item \textbf{CoTAPI-S: } CoT with a pre-defined function set and structuring of the data
\end{itemize}

The techniques involving semi-structured data (data in natural language) are Direct, CoT, ToT, CoTR, CoTE-S \& CoTAPI-S. The techniques that do have access to the already structured data in graph format are CoTE \& CoTAPI. 

For the experiments, we used an AWS machine (t3.xlarge) with 4 vCPUs and 16GB of RAM. The LLM selected for this study is the GPT-4o model ~\cite{openai2024gpt4o} in its "gpt-4o-2024-05-13" version, accessed via the OpenAI API, as it is one of the most advanced LLMs available at the time of experimentation~\cite{lmsys_chatbot_arena_leaderboard}. The prompts used for each technique can  be found in the supplementary material (see ~\ref{prompts})


\subsection{Experiment results}

For the first evaluation, Table~\ref{tab:accuracy} presents the accuracy for each technique along with the average time taken to complete the tasks in the RATA benchmark. The specific results for the Test-of-Time benchmark are available in the Supplementary Material (Table~\ref{tab:accuracy_tot}). Regarding the accuracy, we observe that traditional techniques where the LLM is responsible for performing the entire reasoning process (Direct, CoT \& ToT) are among the ones with the worst results. Additionally, the techniques involving data structuring yield lower accuracy compared to their counterparts. Even the CoTE-S technique achieves lower performance than CoT, indicating the LLM struggles structuring the data (with an average of 3898 tokens per TKG). The CoTAPI technique achieves the best results, outperforming CoTE and CoTR, which obtains surprising results without actual execution of code, showing that reflexion techniques really help during reasoning. In terms of time, the two techniques involving data structuring take the longest time being an order of magnitude slower compared to their counterparts. Moreover, the applicability of the techniques has been proved in another benchmark (CronQuestion) which information and discussion can be found in the supplementary material (Section~\ref{aplicability}).

\begin{table}[h!]
\centering
\small
\begin{tabular}{lrrr}
\toprule
  Method &  Accuracy &  Average Time &  Time (Std Dev) \\
\midrule
  Direct &      59\% &          2.65 &            1.89 \\
     CoT &      64\% &         44.29 &           56.26 \\
     ToT &      57\% &         60.36 &          226.31 \\
     CoTR &      70\% &         36.03 &          176.07 \\
  CoTE-S &      63\% &        109.87 &          250.91 \\
    CoTE &      72\% &         10.27 &           13.74 \\
CoTAPI-S &      82\% &        122.32 &          293.84 \\
  CoTAPI &      \textbf{93\%} &          9.70 &            5.09 \\
\bottomrule
\end{tabular}
\caption{Accuracy and Time values for the different techniques.}
\label{tab:accuracy}
\end{table}

\begin{figure}[h!]
    \centering
    \includegraphics[width=0.49\textwidth]{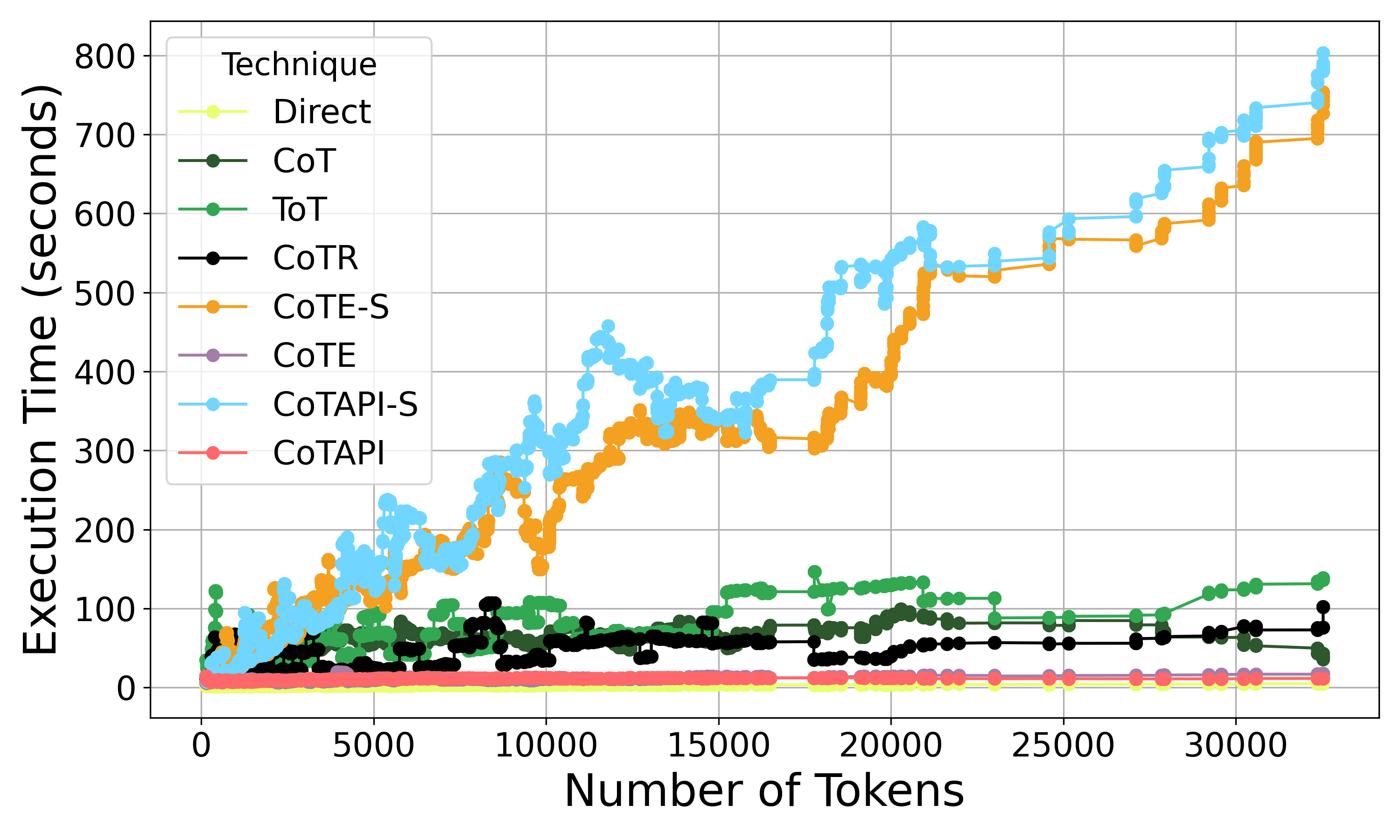}
    \caption{Execution time vs number of tokens.}
    \label{fig:tokensTime}
\end{figure}

Delving into the differences in execution times, Figure~\ref{fig:tokensTime} illustrates the execution time of the techniques as the number of tokens increases (directly related to data size). We observe that time increases with the size of the data in techniques requiring data structuring (CoTE-S \& CoTAPI-S). Conversely, the execution time remains mostly constant in techniques where the data is already structured.

\begin{figure*}[htbp]
    \centering
    \includegraphics[width=0.7\textwidth]{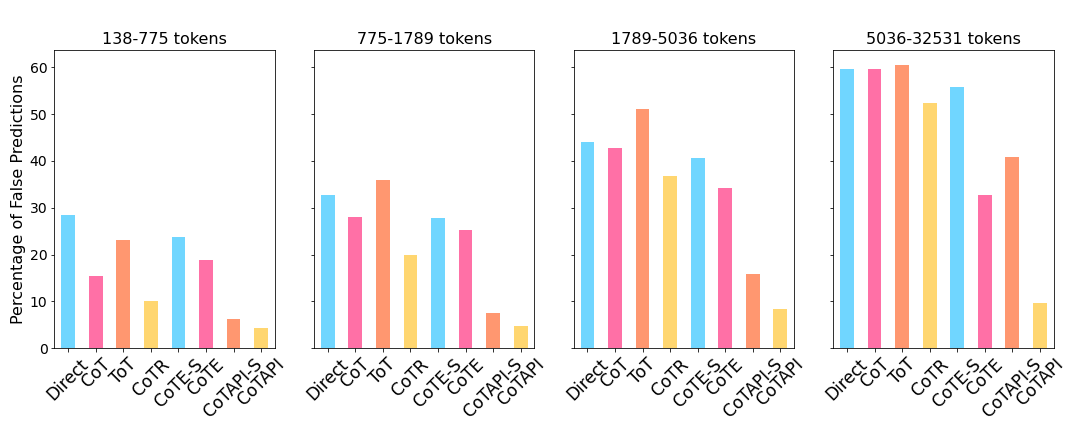}
    \caption{Percentage of false predictions per technique as the size of the data (tokens) increases.}
    \label{fig:accuracyvstokens}
\end{figure*}

Regarding the analysis of the accuracy by question type (Table~\ref{tab:accuracy_by_question}), we see that for certain types of questions that are particularly challenging for the Direct approach (e.g., "Sort entities"), the accuracy improves drastically when a technique involving algorithmic execution is applied. In other scenarios, such as "In between entities", we observe that the CoTE technique does not improve the results and even worsens them, whereas the use of deterministic functions (CoTAPI) enhances accuracy. In Figure~\ref{fig:accuracyvstokens} we see that most techniques tend to increase in its percentage of false predictions as the size of the data increases while CoTAPI remains mostly constant, followed by CoTE.

\begin{table*}[h]
\centering
\small
\begin{tabular}{lrrrrrrrrr}
\toprule
                     Task &  Questions & Direct & CoT & ToT & CoTR & CoTE-S & CoTE & CoTAPI-S & CoTAPI \\
\midrule
                      before\_after &                  350 &    54\% & 57\% & 63\% &  69\% &    43\% &  57\% &      88\% &   \textbf{91\%} \\
     calculate\_total\_relation\_time &                  340 &    50\% & 51\% & 52\% &  39\% &    39\% &  62\% &      85\% &    \textbf{97\%} \\
   check\_interval\_without\_relation &                  340 &    68\% & 78\% & 46\% &  88\% &    56\% &  61\% &      89\% &    \textbf{99\%} \\
         compare\_triplet\_durations &                  340 &    76\% & 94\% & 95\% &  95\% &    79\% &  58\% &      87\% &   \textbf{97\%} \\
     count\_relations\_with\_duration &                  340 &    42\% & 42\% & 45\% &  66\% &    59\% &  89\% &      80\% &   \textbf{99\%} \\
event\_at\_the\_time\_of\_another\_event &                  350 &    71\% & 70\% & 66\% &  80\% &    70\% &  76\% &      77\% &    \textbf{85\%} \\
                   event\_at\_time\_t &                  350 &    69\% & 65\% & 67\% &  83\% &    74\% &  85\% &      90\% &     \textbf{95\%} \\
                event\_at\_what\_time &                  350 &    87\% & 93\% & 94\% &  94\% &    87\% &  94\% &      94\% &    \textbf{99\%} \\
      find\_entities\_during\_triplet &                  340 &    36\% & 35\% & 32\% &  42\% &    44\% &  60\% &      61\% &    \textbf{82\%} \\
                        first\_last &                  350 &    71\% & 72\% & 71\% &  75\% &    79\% & \textbf{89\%} &      80\% &    83\% \\
           get\_entities\_in\_between &                  340 &    12\% &  6\% &  9\% &   6\% &     5\% &   3\% &      52\% &   \textbf{75\%} \\
            get\_entity\_by\_duration &                  340 &    56\% & 66\% & 54\% &  68\% &    68\% &  82\% &      87\% &    \textbf{95\%} \\
        is\_triplet\_within\_timespan &                  340 &    70\% & 92\% & 55\% &  94\% &    81\% &  69\% &      89\% &    \textbf{99\%} \\
 number\_of\_events\_in\_time\_interval &                  350 &    57\% & 56\% & 62\% &  70\% &    84\% &  95\% &      87\% &    \textbf{97\%} \\
                 relation\_duration &                  350 &    83\% & 89\% & 90\% &  92\% &    78\% &  92\% &      91\% &    \textbf{97\%} \\
 sequence\_of\_relations\_in\_interval &                  340 &    74\% & 84\% & 40\% &  84\% &    64\% &  68\% &      81\% &    \textbf{98\%} \\
                          timeline &                  350 &    23\% & 30\% & 32\% &  45\% &    61\% &  86\% &      81\% &    \textbf{95\%} \\
                    Total Accuracy &                 5860 &    59\% & 64\% & 57\% &  70\% &    63\% &  72\% &      82\% &    \textbf{93\%} \\
\bottomrule
\end{tabular}
\caption{Accuracy per type of question for the different techniques.}
\label{tab:accuracy_by_question}
\end{table*}

Another important aspect to analyze is the use of the function set by the techniques that involve it (CoTAPI \& CoTAPI-S).  Table~\ref{tab:use_function_set} presents this analysis comparing the use of the function designed for each question type with the evaluation outcome. We observe that in most cases where a non-associated function is selected, the correct result is still achieved as the LLM has find a turnaround to obtain the correct answer following a different logic path. On the other hand, cases where the associated function is chosen but the prediction is incorrect are due to the LLM failing to correctly identify the parameters of the function. This is due to a hallucination by the model, or a lack of clear semantics in the question.

\begin{table}[h!]
\centering
\footnotesize
\begin{tabular}{lrr}
\toprule
{} &  Asociated Funct. &  Non-Asociated Funct. \\
\midrule
True Pred.  &              85.6\% &                   7.5\% \\
False Pred. &               3.6\% &                   3.2\% \\
\bottomrule
\end{tabular}
\caption{Percentage of use of the designated function vs the evaluation of the prediction.}
\label{tab:use_function_set}
\end{table}

\subsubsection*{Major improvement of using external tools}

By conducting a more exhaustive analysis between the Direct technique and the technique with the best results, CoTAPI, we can derive some insights into the limitations of current LLMs in temporal reasoning, and the advantages gained from using external deterministic tools, leading to an improvement in accuracy from \textbf{59\%} to \textbf{93\%} while maintaining the same order of magnitude in execution time.

\begin{figure}[ht]
    \centering
    \includegraphics[width=0.4\textwidth]{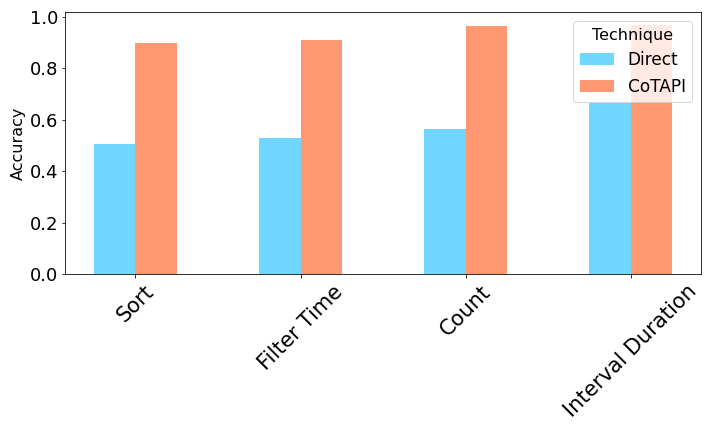}
    \caption{Accuracy grouped by algorithms.}
    \label{fig:accuracy_by_algorithm}
\end{figure}

Taking a look at the accuracy values grouped by the different main algorithms that compose the tasks as identified in Section \ref{temporal_tasks} (Figure \ref{fig:accuracy_by_algorithm}), we observe that the "Sort" algorithm is the most challenging one for the LLM, while the "Interval Duration" is the easiest. This result correlates with the theory that, when the LLM requires a complex algorithm like a sorting algorithm, it will perform worst than when just requiring a simpler algorithm like filtering based of the duration of an event. In contrast, the CoTAPI technique achieves very good results across all these algorithms, as the LLM task is agnostic of the internal algorithm required, and focuses solely on the semantic analysis.

\begin{figure}[ht]
    \centering
    \includegraphics[width=0.4\textwidth]{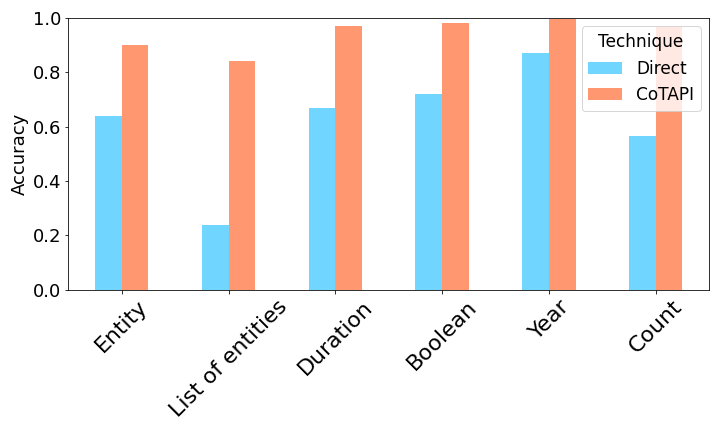}
    \caption{Accuracy grouped by the response type.}
    \label{fig:response_type}
\end{figure}

Moving on to the accuracy grouped by the response type (Figure \ref{fig:response_type}), we observe a huge drop in accuracy when the response is a list of entities, indicating that the LLM has difficulties when it needs to identify multiple entities as a response rather than a single piece of data. While this multi-entity task can be straightforward in a logical algorithm, it becomes more challenging when relying solely on semantic analysis of the text to find several entities in the correct order.

%% file: Sections/8.aplicability.tex
\subsection{Applicability} \label{aplicability}
The compared methods can be adapted and applied to many other scenarios beyond the RATA dataset used for evaluation. To show so, we have evaluated also the CronQuestions dataset~\cite{cronquestions} in its anonymized version and compared against state-of-the-art (SoTA) techniques. This dataset was not originally designed for being used with LLMs. Hence, it makes it a good example for testing the effectiveness of the methods, but also presents some limitations. Specifically, only the methods involving external structured data access (CoTE \& CoTAPI) can be tested, as the size of the TKG in a semi-structured natural language format in this dataset exceeds the context window of the selected LLM (128K tokens). Furthermore, we have observed that the data structure used in this dataset resembles the triplet format used in our RATA dataset but also has scenarios where the triplet is used to specify explicit time information such as \textit{("Reign of Terror", "significant event", "occurrence", 1793, 1794)}, which does not resemble a relationship between two entities, but the duration of a specific event. As this format was not included during evaluation, we anticipate poorer results.

Results in Table~\ref{tab:cronquestions} indicate that both techniques could be used in other scenarios. The SoTA techniques that surpass in accuracy the CoTE \& CoTAPI techniques are those which require some kind of training. Regardless, this show how LLMs can compete against more specific tailored techniques only using prompting and external execution techniques. Moreover, the SoTA techniques do not generalize to other scenarios as well as the compared LLM-techniques, as these are specifically tailored to the data structure of the CronQuestion dataset, only cover the type of the answer being an entity (not a list of entities, duration, boolean or count) or require the semantic information and general knowledge to obtain the answer (not valid for anonymized scenarios), making them not suitable for more general and different scenarios, like the proposed RATA dataset \cite{mavromatis2022tempoqr, chen2024selfimprovementprogrammingtemporalknowledge, chen-etal-2024-temporal, liu2023timeawaremultiwayadaptivefusion}. This show us how capable are LLMs with external execution in generalization even in the TQA task without training involved, but still, a specifically tailored technique can obtain better results using smaller and more-optimized models.

\begin{table}[h!]
\centering
\small
\begin{tabular}{lr}
\toprule
Method &  Accuracy \\
\midrule
T5-3B &      7.3\% \\
BERT &      8.6\% \\
EmbedKGQA &      28.6\% \\
CronKGQA &      39.2\% \\
ARI &      57.0\% \\
TMA &      63.2\% \\
TempoQR-Soft* &      65.5\% \\
TempoQR-Hard* &      86.4\% \\
Prog-TQA* &      89.8\% \\
CoTE &      47.9\% \\
CoTAPI &      67.0\% \\
CoTAPI-Adapted &      74.5\% \\
\bottomrule
\end{tabular}
\caption{Results over the CronQuestions dataset Hits@1. \textit{*Requires training}}
\label{tab:cronquestions}
\end{table}

%% file: Sections/9.temporal_condifence.tex
\subsection{Temporal Confidence}
One of the key applications is the development of a temporal-expert LLM agent. To do this, the agent needs to determine whether the QA task involves any temporal reasoning, and if so, apply the appropriate technique. To simulate this scenario we developed a prompt through which the LLM analyzes the task along with the structure of the data, and estimates a \textit{temporal confidence} value between 0 and 1. Values close to 0 indicate that the task does not involve temporal reasoning, while values close to 1 indicate the task is related to TQA.

To evaluate this technique, we combined the excerpt of 1,000 KQA tasks from the dataset ComplexWebQuestions \cite{talmor2018webknowledgebaseansweringcomplex} with 1,000 TQA tasks from our RATA dataset. The proposed temporal confidence estimation technique was tested using the GPT-4o model with a threshold of 0.8 over 1 to identify TQA tasks, achieving an accuracy of \textbf{97\%}. This result validates the feasibility of developing an agent capable of distinguishing TQA tasks from KQA for a proper treatment.

\begin{table}[h!]
\centering
\small
\begin{tabular}{lrr}
\toprule
Predicted &  Knowledge &  Temporal \\
Actual    &            &           \\
\midrule
Knowledge &        940 &        60 \\
Temporal  &          0 &      1000 \\
\bottomrule
\end{tabular}

\caption{Confusion matrix of the predicted vs actual values of the temporal classification.}
\end{table}

%% file: Sections/10.discussion.tex
\section{Discussion}

Looking at the results, we realized that the standalone capabilities of LLMs in the TQA task over semi-structured anonymized data are not sufficient. While they can sometimes identify correct answers when the TKG is small, more complex scenarios (like real-world) fall outside of their capabilities (See Figure~\ref{fig:accuracyvstokens}). These tasks require different operations and algorithms to be properly resolved like sequencing of events, arithmetic computation, or temporal constraint checking, and mere semantic analysis of the data, as performed by the LLM during inference, is not enough to solve them.

The CoT method shows little improvement as it tends to be effective when tasks can be pre-divided into distinct steps. In this scenario, the tasks are varied and unpredictable beforehand (as real-world), making such pre-division not feasible. Additionally, these steps would still require solving tasks that the LLM struggles with, which relate to the little improvement achieved with the ToT method. With the CoTR method we observe improvement showing the effectiveness of letting the LLM revise its own answer and trying to improve.

When testing the LLM's ability to structure semi-structured data in conjunction with temporal reasoning, we observe they yield worse results compared to their counterparts where the data are already structured, and even worst results compared to techniques not involving external executions. It is also relevant to notice that the percentage of erroneous predictions increases with the size of the data more heavily than compared to the techniques not involving structuring the data (Figure~\ref{fig:accuracyvstokens}), demonstrating how the structuring capability of the LLM degrades overall performance when coupled with other task.

Focusing now on the two most promising proposed methods (CoTE \& CoTAPI), these improve the baseline by a significant margin. CoTE resembles the most attractive option when the task set is not known, while CoTAPI emerges as the superior option when looking for reliability and accuracy. This suggests that achieving a scalable and reliable solution requires combining both methodologies while relieving the LLM from the actual reasoning, achieving more reliability when the type of question has been previously recognized but also being able to scale to less common new problems.

%% file: Sections/11.conclusion.tex
\section{Conclusion}

During this study, we conducted an analysis of TQA tasks, identifying the 17 most common types and performing an in-depth examination of their algorithmic resolution. To support the evaluation, a TQA dataset LLM-focused called RATA was developed consisting of these 17 question types, and encompassing a total of 5,850 questions built upon semi-structured anonymized data preventing the LLM from relying on prior knowledge. We compared several methodologies involving LLM for structured and semi-structured data, where two of them stood out from the others (CoTE \& CoTAPI), being the first one a suitable option for zero-shoot scenarios and the second one for few-shot scenarios, achieving 72\% \& 93\% of accuracy compared to the baseline LLM performance of 59\% while maintaining both techniques bounded execution time, remarking the usefulness of using external execution for logic-based reasoning. In different scenarios these techniques do not beat specifically tailored SoTA techniques, but can compete against them without training or tuning required, showing a remarkable generalization capability. We have also successfully checked the temporal confidence of the LLM which enables its potential application in a temporal-expert LLM agent.



%% file: Sections/13.limitations.tex
\section{Limitations}

While this study provides valuable insights into the temporal reasoning capabilities of Large Language Models (LLMs), and proposes effective methodologies for enhancing these capabilities, several limitations must be acknowledged.

The experiments in this study were conducted using only one LLM, specifically the GPT-4o model in its "gpt-4o-2024-05-13" version. Although, at the time of experimentation, this model was recognized as the one obtaining the best results across different aspects by the community~\cite{lmsys_chatbot_arena_leaderboard}, the results may not generalize across other LLMs for this specific experimentation scenario. Future research should consider evaluating multiple LLMs to provide a more comprehensive understanding of their temporal reasoning capabilities.

The study was conducted exclusively in English, which may limit the generalizability of the findings to other languages. In that case, the LLM might have lower capabilities as the corpus of training data for these models is not balanced, and have more quantity of English-related data than other types of languages. Moreover, the field of LLMs and AI, in general, is rapidly evolving. New models with enhanced capabilities are frequently being developed and released. As a result, the findings of this study may become outdated as newer models with improved temporal reasoning abilities emerge. Continuous evaluation and adaptation of the proposed methodologies are necessary to keep pace with advancements in the field.

Lastly, focusing on the elaborated \textit{Reasoning and Answering Temporal Ability} dataset (RATA) and the CoTAPI technique, these were based and experimented on synthetic data composed of a limited set of types of TQA tasks. This set may not cover all possible types of TQA tasks, and the complexity and nuances of real-world data. Future research should consider trying out real-world TQA problems.
An example of this limitation was encountered when applying the technique of CoTAPI to the CronQuestions dataset (Section~\ref{aplicability}). After a thorough analysis of the different question types in this dataset, and how each technique performed, two scenarios were identified that were not directly covered by the existing function set. These scenarios involved implicit time references in a unique way, where the Temporal Knowledge Graph (TKG) not only stores triplets, but also a single entity with a time reference. As a result, the function set would need to be adapted to this specific data structure to achieve even better accuracy results, which differs from the one commonly found in the literature. 

By acknowledging these limitations, we aim to provide a balanced perspective on the findings of this study, and highlight areas for future research and improvement.



%% file: Sections/12.disclaimer.tex
\section*{Disclaimer}
This paper was prepared for informational purposes by the Artificial Intelligence Research group of JPMorgan Chase \& Co. and its affiliates ("JP Morgan'') and is not a product of the Research Department of JP Morgan. JP Morgan makes no representation and warranty whatsoever and disclaims all liability, for the completeness, accuracy or reliability of the information contained herein. This document is not intended as investment research or investment advice, or a recommendation, offer or solicitation for the purchase or sale of any security, financial instrument, financial product or service, or to be used in any way for evaluating the merits of participating in any transaction, and shall not constitute a solicitation under any jurisdiction or to any person, if such solicitation under such jurisdiction or to such person would be unlawful.

© 2025 JPMorgan Chase \& Co. All rights reserved

%% file: Sections/14.prompts.tex
\subsection{Prompts} \label{prompts}
Next we include the full list of prompts used during experimentation. Each of them has been developed in an iterative manner with trial and error but following the guidelines of the prompt engineering philosophy and looking at the papers of each of the techniques used in each methodology.

\subsubsection{Chain of Thought}

\textbf{Prompt 1}: \textit{Next I am going to give you a set of data and a task to work on. First I want you to think about the task you are asked to perform and the nature of the data. Return a brief paragraph about how the data is described and then another paragraph about how do you think would be the most optimal way of reason and represent the task. Then, extract the task to be performed and the structure of the desired result and output it between $<start_task>$ and $<end_task>$. Do not solve the task in the statement, just analyze it. Output the full answer. The statement is the following: $<start_statement> {task} <end_statement>$. Focus on the first task about analyzing the data and task.}

\textbf{Prompt 2}: \textit{Following the chain of thought philosophy, first focus on the data. Identify, extract and represent the most relevant data in the most optimal way so that later, when you will carry out the designated task, it will be solved in the simplest and most optimal way possible. Pay attention to the nature of the task and the data to know in which structure to represent them. Output the full answer and do not short the answer for brevity. Do not solve the main task now, just identify and extract the relevant data}

\textbf{Prompt 3}: \textit{With the statement parsed and the data structured in the format you have chosen, focus now only on solving the task you have been given within the statement. You must return only the response in JSON format as indicated in the task of the initial statement with the fields 'explanation' and 'answer' as requested in the statement.}

\subsubsection{Tree of Thought}

\textbf{Initial Prompt}: \textit{You are an expert in temporal question-answering problems. Next I am gonna give you a task of temporal question-answering and semi-structured data over which you have to reason. I want to follow the tree of thought philosophy, where you would have to reason step by step over the data and the task to obtain insights required to solve the task in ${max_depth}$ different levels (or steps) and at the end you would have to obtain the solution based on the best logical path. For each level of reasoning, you would have to get different insights and propose different ideas base on the data and the task. Later on, you will be also asked to select the most appropriate ones.}

\textbf{Expand node prompt}: \textit{You are solving step by step a temporal question-answering problem based on a semi-structured data. This is the ${depth}$ level of reasoning, up to ${max_depth}$ levels in total before trying to obtain the answer: Task and data: ${task}$. Question: ${question}$. Your current reasoning state is:${node_state}$. Based on the current status, Please provide three different insights and possible next reasoning steps over the data and task suitable to solve the task successfully. Do not solve the task directly, think about next possible logical steps to solve it instead. Return the steps in your JSON format ($```json ```$) with the keys "Option1", "Option2" and "Option3", each containing only a string with the next reasoning step thought and the insights}

\textbf{Evaluate State prompt}: \textit{Evaluate the following reasoning state for the given temporal question-answering problem based on a semi-structured data. Task and data: ${task}$. Reasoning state: ${state}$. Provide a score from 1 to 10, where 10 indicates excellent reasoning towards the solution, and 1 indicates poor or incorrect reasoning. Return the only the answer in your JSON format ($```json ```$) with the field "score"}

\textbf{Final Prompt}: \textit{Based on the following reasoning, provide the final answer to the temporal question-answering question. Task and data: ${task}$. Final reasoning path: ${reasoning_path}$. Provide only the answer in you JSON format ($```json ```$) with a single attribute "answer" containing only the answer to the question.}

\subsubsection{Chain of Thought with Self-Reflexion}

\textbf{Prompt 1}: \textit{Next I am going to give you a set of data and a task to work on. First I want you to think about the task you are asked to perform and the nature of the data. Return a brief paragraph about how the data is described and then another paragraph about how do you think would be the most optimal way of reason and represent the task. Then, extract the task to be performed and the structure of the desired result and output it between $<start_task> and <end_task>$. Do not solve the task in the statement, just analyze it. Output the full answer. The statement is the following: $<start_statement> {task} <end_statement>$. Focus on the first task about analyzing the data and task.}

\textbf{Prompt 2}: \textit{Following the chain of thought philosophy, first focus on the data. Identify, extract and represent the most relevant data in the most optimal way so that later, when you will carry out the designated task, it will be solved in the simplest and most optimal way possible. Pay attention to the nature of the task and the data to know in which structure to represent them. Output the full answer and do not short the answer for brevity. Do not solve the main task now, just identify and extract the relevant data}

\textbf{Prompt 3}: \textit{With the statement parsed and the relevant data identified, focus now only on solving the task you have been given within the statement. You must return only the response in JSON format as indicated in the task of the initial statement with the fields 'explanation' and 'answer' as requested in the statement.}

\textbf{Reflexion Prompt}: \textit{Analyze the result you have obtained based on the task and the data you have identified. If it seems as a possible correct answer, return the answer again in the json formate requested between your $```json ```$ identifiers. If not, return the term $"repeat_please"$ and the reasoning why you believe the answer you have obtained is not correct, which reasoning could help you to obtain the correct one in a future iteration}

\subsubsection{Chain of Thought with External Execution}

\textbf{Prompt 1}: \textit{Next I am going to give you a set of data and a task to perform on it. First I want you to think about the task you are asked to perform and the nature of the data. Return a brief paragraph about how you think you would solve the task using graphs. Then, extract the task to be performed and the structure of the desired result and output it between $<start_task> and <end_task>$. Do not solve the task in the statement, just analyze it. Output the full answer. The statement is the following: $<start_statement> {task} <end_statement>$. Focus on the first task about thinking and structuring the statement.}

\textbf{Prompt 2 (Applicable for structuring the data for CoTE-S, letting the LLM decide the best way for representing the data)}: \textit{Now, take the data and transform it into a temporal graph attending to the nature of the task to be performed later. To do this, make an untagged, unannotated executable code that transforms that data into a graph named "G" using the NetworkX library in python. First define the graph G and then add the edges one by one. The code should contain only executable code, no comments nor annotations. Note that between two nodes there can be more than one edge. Write the complete code with all the necessary data to just take it, run it and get the graph. Do not do any print or debug. Do not generate any kind of explanation or extra information other than what I have asked for. Output the full answer code in python code format, do NOT short the answer for brevity and focus on not missing any row in the original order and not repeating the same data.}

\textbf{Prompt 3}: \textit{Now, knowing the data, its format and that the previously data is already defined in a MultiDiGraph of NetworkX in python with the name of "G" and that the data has been inserted using sentences like: $"G.add_edge("E12", "E45", relation="R12", start_time=1975, end_time=2000)"$, make the code executable to get the answer to the task of the statement. Do not generate extra information, just the executable code to solve the task. The code should contain only executable code, no comments nor annotations. Do not make any method or function, directly the code to solve the task. Do not output the previously generated code about the generation of the graph, just focus on solving the task. Do not generate any extra information, just the code in python code format to perform the task assuming the graph is already defined. Instead of output the JSON, save it in a variable called 'result' in a dict structure with the fields 'explanation' and 'answer' as requested in the statement.}

\subsubsection{Chain of Thought with API Access}

\textbf{Prompt 1}: \textit{Next I am going to give you a set of data and a task to perform on it. First I want you to think about the task you are asked to perform and the nature of the data. Return a brief paragraph about how you think you would solve the task using graphs. Then, extract the task to be performed and the structure of the desired result and output it between $<start_task> and <end_task>$. Do not solve the task in the statement, just analyze it. Output the full answer. The statement is the following: $<start_statement> {task} <end_statement>$. Focus on the first task about thinking and structuring the statement.}

\textbf{Prompt 2 (Applicable for structuring the data for CoTAPI)}: \textit{Now, take the data and transform it into a temporal graph attending to the nature of the task to be performed later. To do this, make an untagged, unannotated executable code that transforms that data into a graph named "G" using the NetworkX library in python. First define the graph G as a MultiDiGraph and then add the edges one by one. Use a structure similar to this one: $G.add_edge("XXX", "XXX", relation="XXX", start_time=XXXX, end_time=XXXX)$. The code should contain only executable code, no comments nor annotations. Note that between two nodes there can be more than one edge. Write the complete code with all the necessary data to just take it, run it and get the graph. Do not do any print or debug. Do not generate any kind of explanation or extra information other than what I have asked for. Output the full answer code in python code format, do NOT short the answer for brevity and focus on not missing any row in the original order and not repeating the same data.}

\textbf{Prompt 3}: \textit{Now, knowing the data, its format and that the previously code you generated was already executed, use the functions I have given to you to get the answer to the task of the statement. Focus on the task to think which function or functions will give you the most suitable response to avoid using irrelevant functions for the task. The task can be resolve by using only one function, so focus which is the most suitable. Do not generate comments, just use the functions to obtain relevant insights for resolving the task. In the next iteration you will make use of the results from these functions you selected to obtain the final result.}

\textbf{Prompt 4}: \textit{With the results of the functions you have selected, make use of this results to resolve the task described previously and output the answer in the JSON format indicated at the beginning in the statement with the fields 'explanation' and 'answer' as requested in the statement. Not plain text neither executable code, output it between your JSON labels $```json ```$}

%% file: Sections/15.functions.tex
\subsection{Function structure example}
We show now a piece of the function structure provided to the LLM to give it some context of the available functions, when to use them and the parameters required by each of them. Each of the function description and logical code associated were developed focusing on each of the question types previously identified in Table~\ref{tab:question_types}. The code was developed looking at their low-level algorithms and their description with possible use cases in mind.
\onecolumn
\begin{lstlisting}[language=json]
[
    {
        "type": "function",
        "function": {
            "name": "get_relations_node",
            "description": "Use this function to getting a list with all the nodes, relations and atributes associated to a specific node.",
            "parameters": {
                "type": "object",
                "properties": {
                    "node": {
                        "type": "string",
                        "description": "Name of the specific node for which to get all its neighbours with relation and atributes."
                    }
                },
                "required": [
                    "node"
                ]
            }
        }
    },
    {
        "type": "function",
        "function": {
            "name": "before_after",
            "description": "Use this function when knowing a node origin, a relation and a destination node, you want to find the previous or next origin node chronologically of that relation with the destination node. An example use case could be: Which entity was the RXX of the entity EXX immediately after EXX?",
            "parameters": {
                "type": "object",
                "properties": {
                    "node_origin": {
                        "type": "string",
                        "description": "Name of the specific origin node for which to get the previous or next chronologically node."
                    },
                    "relation": {
                        "type": "string",
                        "description": "Name of the specific constant relation between the desired origin node and destination node."
                    },
                    "node_destination": {
                        "type": "string",
                        "description": "Name of the specific destination constant node."
                    },
                    "objective": {
                        "type": "string",
                        "description": "\"next\" if you are looking for the next node chronologically or \"previous\" if looking for the previous one"
                    }
                },
                "required": [
                    "node_origin",
                    "relation",
                    "node_destination",
                    "objective"
                ]
            }
        }
    },
    {
        "type": "function",
        "function": {
            "name": "timeline",
            "description": "Use this function to get a list sorted with all the origines node that have had a relation known with a specific destination node. An example use case could be: Which entities were the RXX of EXX sorted by end time from lowest to highest?",
            "parameters": {
                "type": "object",
                "properties": {
                    "node_destination": {
                        "type": "string",
                        "description": "Name of the specific destination node for which to get the list of all the origin nodes with a specific relation between them."
                    },
                    "relation": {
                        "type": "string",
                        "description": "Name of the specific relation between the specific destination node and all the origin nodes to be returned as a list"
                    },
                    "sorted_key": {
                        "type": "string",
                        "description": "Name of the key for which to sort by. Could be \"start_time\" or \"end_time\"."
                    },
                    "sorted_direction": {
                        "type": "string",
                        "description": "\"increasing\" if the sort goes from the lowest to the highest or \"decreasing\" if reverse."
                    }
                },
                "required": [
                    "node_destination",
                    "relation",
                    "sorted_key",
                    "sorted_direction"
                ]
            }
        }
    }
]
\end{lstlisting}